\documentclass{article}

\usepackage[preprint]{neurips_2025}

\usepackage[utf8]{inputenc} % allow utf-8 input
\usepackage[T1]{fontenc}    % use 8-bit T1 fonts
\usepackage{hyperref}       % hyperlinks
\usepackage{url}            % simple URL typesetting
\usepackage{booktabs}       % professional-quality tables
\usepackage{amsfonts}       % blackboard math symbols
\usepackage{nicefrac}       % compact symbols for 1/2, etc.
\usepackage{microtype}      % microtypography
\usepackage{xcolor}         % colors

\usepackage{graphicx}
\usepackage{multirow}
\usepackage{array}
\usepackage{amsmath, amssymb}
\usepackage{siunitx}

\title{A Framework for Nonstationary Gaussian Processes with Neural Network Parameters}

\author{%
  Zachary James \\
  Department of Statistics and Data Science\\
  Cornell University\\
  Ithaca, NY 14850 \\
  \texttt{zj37@cornell.edu} \\
  \And
  Joseph Guinness \\
  Department of Statistics and Data Science\\
  Washington University\\
  St Louis, MO 63130 \\
  \texttt{joeguinness@wustl.edu} \\
}
\usepackage[normalem]{ulem}

\begin{document}

\maketitle

\begin{abstract}
  Gaussian processes have become a popular tool for nonparametric regression because of their flexibility and uncertainty quantification. However, they often use stationary kernels, which limit the expressiveness of the model and may be unsuitable for many datasets. We propose a framework that uses nonstationary kernels whose parameters vary across the feature space, modeling these parameters as the output of a neural network that takes the features as input. The neural network and Gaussian process are trained jointly using the chain rule to calculate derivatives. Our method clearly describes the behavior of the nonstationary parameters and is compatible with approximation methods for scaling to large datasets. It is flexible and easily adapts to different nonstationary kernels without needing to redesign the optimization procedure. Our methods are implemented with the GPyTorch library and can be readily modified. We test a nonstationary variance and noise variant of our method on several machine learning datasets and find that it achieves better accuracy and log-score than both a stationary model and a hierarchical model approximated with variational inference. Similar results are observed for a model with only nonstationary variance. We also demonstrate our approach's ability to recover the nonstationary parameters of a spatial dataset. 
\end{abstract}

\section{Introduction}

Gaussian processes (GPs) use distributional assumptions to infer the value of an unknown function at different input values. These assumptions result in a model that is both highly interpretable, with prediction intervals that can be easily computed, and flexible, as the assumptions themselves can incorporate domain knowledge. As a result, GPs have become popular in nonparametric regression \cite{rasmussen04}, spatial statistics \cite{campsvalls16}, and computer emulation \cite{qian08}.

However, GP models often assume that the data generating process has a covariance function, also called the kernel, that is either stationary (a function of the difference between feature vectors) or isotropic (a function of the Euclidean distance between feature vectors). Such assumptions are often made without considering the data and may be unjustified \cite{sampson92}. Many processes have variances, lengthscales, noise, or anisotropy that vary throughout the feature space.  

A large number of nonstationary GP methods have been developed to address this. Warping methods transform the feature space such that the resulting process is stationary \cite{sampson92}. Possible transformations include splines \cite{anderes08}, compositions of functions \cite{zammit22}, and neural networks \cite{wilson16}. It is often desired that the transformation be an injection, and several methods explicitly enforce this constraint \cite{anderes08, wilson16}. Similar to warping methods, Deep GPs model a GP on a transformed space. The transformation is expressed as the composition of latent functions, with GP priors placed on each one \cite{damianou13}. Since computing parameter estimates directly is intractable, variational inference is often used instead.

Partition based methods fit separate models to different regions. These regions may be decided by using domain knowledge, learned \cite{gramacy08, muyskens22}, or the result of a sliding window \cite{gerber2021}. You can also learn separate models for each prediction point \cite{haas95, gramacy14}.

Another approach is to use a nonstationary covariance function with parameters that vary across the feature space. \cite{paciorek06} proposed a covariance function with varying anisotropy and then placed a GP prior on the parameter. This has been similarly done with the noise \cite{goldberg97} and lengthscale \cite{heinonen16}, with Markov chain Monte Carlo (MCMC) sampling used for inference in each case. In spatial statistics, considerable work has focused on expressing the process as a linear combination of coefficients and basis functions \cite{cressie22}. Nonstationarity can be captured by letting the coefficients or functions vary as a function of the input. Alternatively, a nonstationary covariance function can be used with the nonstationary parameters expressed as a function of basis functions \cite{blasi2024}.

While effective, several of these methods may be difficult to implement or be unable to model certain forms of nonstationarity. Warping methods are computationally efficient, but cannot capture nonstationary variance or nonstationary noise. Nonstationary covariance functions with parameters that vary across the feature space are highly interpretable, but implementation is challenging. Specifying the models usually involves expressing the nonstationary parameters as linear combinations of basis functions, whose definition may be problem-dependent. Model-fitting is also difficult, whether optimizing the likelihood over a high-dimensional parameter space, or sampling from a high-dimensional posterior, especially when there are multiple nonstationary parameters \cite{heinonen16}. 

We propose a framework that allows for different nonstationary kernels to be easily implemented. We model the nonstationary parameters in a nonstationary covariance function as the output of a feed-forward neural network that takes the features as input. The GP and neural network are trained together, with automatic differentiation used to calculate derivatives. This method can be interpreted as learning basis function expansions for the nonstationary parameters, with the last hidden layer of the neural network as the learned basis functions and the weights of the last layer as the coefficients. Different nonstationary kernels can be used, with only the output layer of the neural network needing to be changed. Furthermore, we find that shallow neural nets are sufficient, minimizing the need for extensive model tuning. We implement all methods with the GPyTorch library \cite{gardner18}, making them easy-to-use.

\section{Gaussian Processes}

Given observations $Y\in\mathbb{R}^{n}$ and features $X\in\mathbb{R}^{n\times d}$, we wish to learn their relationship, which we model as
\begin{equation}
\label{eq:gpmodel}
    y_i = f(x_i) + \epsilon_i
\end{equation}
where $y_i$ is the $i$th element of $Y$, $x_i$ is the $i$th row of $X$, and $\epsilon_i\sim \mathcal{N}(0,\tau_i^2)$ is independent noise. Our goal is to infer the function $f$. To conduct inference we first place a GP prior on the function $f \sim \mathcal{GP}(\mu(\cdot),k(\cdot,\cdot\mid\theta))$, where $\mu(x_i)$ is the expected value of $f(x_i)$, and $k(x_i,x_j \mid \theta)$ is the covariance between $f(x_i)$ and $f(x_j)$. The covariance parameters $\theta$ typically must be learned from the data. 

Given test data $X^* \in \mathbb{R}^{m\times d}$, prediction is performed by deriving the posterior predictive mean and variance of the function
\begin{equation}
\label{eq:predictivemean}
    \mu_{f\mid (X,Y)}(X^*) = \mu(X^*) + K(X^*,X)(K(X,X)+D)^{-1}( Y - \mu(X) )
\end{equation}
\begin{equation}
\label{eq:predictivevar}
    K_{f\mid (X,Y)}(X^*,X^*) = K(X^*,X^*) -K(X^*,X)(K(X,X)+D)^{-1}K(X,X^*)
\end{equation}
where $K(X,X)$ is the covariance matrix for the observations, $K(X,X^*)$ is the cross-covariance matrix, and $K(X^*,X^*)$ is the covariance matrix for the test points. $\mu(X)$ and $\mu(X^*)$ are defined similarly.
The noise matrix is $D = \mbox{diag}(\tau_1^2,\ldots,\tau_n^2)$.

The posterior mean and covariance provide predictions and the associated confidence intervals. The covariance of the prior is parameterized over a class of functions, with the choice of class playing a critical role in performance. Common covariances include the RBF kernel and the Matérn

\begin{equation}
\label{eq:matern}
    k(x,x'\mid\sigma,\rho,\nu) = \sigma^2\frac{2^{1-\nu}}{\Gamma(\nu)}\left(\frac{\sqrt{2\nu}\lVert x-x'\rVert}{\rho}\right)^\nu\mathcal{K}_\nu\left(\frac{\sqrt{2\nu}\lVert x-x'\rVert}{\rho}\right)
\end{equation}
where $\Gamma$ is the gamma function, $\mathcal{K}_\nu$ is the modified Bessel function of the second kind, and $\lVert\cdot\rVert$ is the Euclidean norm. Two frequent assumptions are that the kernel is stationary $k(x,x'\mid\theta)=k^{stat}(x-x'\mid\theta)$ or even isotropic $k(x,x'\mid\theta)=k^{iso}(\lVert x-x'\rVert\mid\theta)$.

\section{Related Work}

\paragraph{Nonstationary Kernels}
A large class of covariance functions express parameters as functions that vary across the feature space. \cite{paciorek06} developed the covariance function

\begin{equation}
\label{eq:paciorek}
    k(x,x'\mid \theta,\sigma(\cdot),\Sigma(\cdot))  = \sigma(x)\sigma(x')|\Sigma(x)|^\frac{1}{4}|\Sigma(x')|^\frac{1}{4}\left\lvert\frac{\Sigma(x)+\Sigma(x')}{2}\right\rvert^{-\frac{1}{2}}k^{iso}\left(\sqrt{Q(x,x')}\mid\theta\right)
\end{equation}
\begin{equation}
    Q(x,x') = (x-x')^T\left(\frac{\Sigma(x)+\Sigma(x')}{2}\right)^{-1}(x-x')
\end{equation}
where $\sigma(\cdot)$ and $\Sigma(\cdot)$ represent local variances and anisotropies and $k^{iso}$ is any isotropic covariance function. Other covariance functions allow for varying noise \cite{goldberg97} or lengthscale \cite{higdon99}. Since the varying parameter $\Tilde{\theta}(\cdot)=(\sigma(\cdot),\Sigma(\cdot))$ is a latent process, estimating it may be challenging. One approach is to place GP priors on the nonstationary parameters and then conduct MCMC sampling, which allows for full posterior inference over the nonstationary parameters and performs very well on small datasets. However, MCMC sampling scales poorly with the number of observations and often needs to be fine tuned depending on the covariance function. Other approaches include expectation-maximization for nonstationary noise models \cite{kersting2007} and variational inference techniques that approximate the posterior \cite{lazaro2011,tolvanen2014, paun2023}.

\paragraph{Basis Functions}
There has been considerable work in the spatial statistics community on expressing spatial processes in terms of basis functions \cite{cressie22}. Given process $y$ observed at location $s$, we express it in terms of its Karhunen–Loève expansion

\begin{equation}
\label{eq:kl}
    y(s) = \sum_{i=1}^\infty \phi_i(s)\lambda_i
\end{equation}
where $\phi_i$ are orthonormal eigenfunctions and $\lambda_i$ are uncorrelated random coefficients. Truncating the series and relaxing the orthogonality produces a basis function approximation. \cite{cressie08} uses a small number of basis functions and then estimates the coefficients, allowing for inference on large datasets. \cite{higdon99} uses a class of spatially varying kernels for the basis functions in order to capture nonstationary behavior. \cite{lindgren11} expressed the process as the solution to a stochastic partial differential equation. Basis functions are then defined on a mesh and finite element methods are used for computational efficiency. 

Many types of basis function have been proposed, such as splines, Wendland functions, Gaussian kernels, and bisquare functions \cite{cressie22}. Since spatial datasets are often low-dimensional using the features themselves as basis functions may not be feasible. However, additional covariates may be introduced and used as basis functions \cite{reich11}.

Basis function approximations have also been used to simplify the task of estimating spatially varying parameters. \cite{paciorek06} expressed the elements in the anisotropy parameter with a basis function approximation to reduce the computational cost of MCMC sampling. If basis functions are predefined for a nonstationary parameter, the coresponding coefficients can be estimated. Several software packages have been developed to fit such models, with \cite{gpgp} providing support for nonstationary variances and \cite{cocons} allowing the user to select from a wide range of nonstationary functions. However, choosing the type and number of basis functions can be a challenge, as they have a direct impact on the performance and computational cost of the model.

\label{sec:deeparam}
\section{Learning Nonstationary Kernel Parameters with Neural Networks}

Our goal is to provide a framework that allows for practitioners to easily fit a large number of different nonstationary kernels. Changing the kernel should not necessitate a completely different optimizer or sampler. Furthermore, we seek to minimize the need for extensive model tuning, such as choosing the type and number of basis functions. Our solution is to use a nonstationary kernel with varying parameters, and then approximate these parameters with a basis function expansion. But rather than predefining the basis functions, we seek to learn them with a neural network. %We minimize the need for model tuning by using shallow neural networks.

\subsection{Model}

Consider nonstationary covariance function $k(\cdot,\cdot\mid\theta, \Tilde{\theta}(\cdot))$ with stationary parameters $\theta$ and nonstationary parameters $\Tilde{\theta}(\cdot)$. We express the nonstationary parameters as a function of a feed-froward neural network $g$ with weight and bias parameters $w$. The neural network is a mapping from the feature space $\mathbb{R}^p$ to the nonstationary parameter space $\Tilde{\Theta}$. In the context of the nonparametric regression problem described in Equation \ref{eq:gpmodel}, we have placed the prior 
\begin{equation}
\label{eq:newmodel}
    f\sim \mathcal{GP}(\mu(\cdot), k(\cdot,\cdot\mid\theta, w) )
\end{equation}
on the function, where $k$ depends on $w$ through $\Tilde{\theta}(\cdot)=g(\cdot\mid w)$. This framework can be used with any nonstationary covariance function that has parameters that vary across the feature space. Consider the nonstationary variance kernel
\begin{equation}
\label{eq:nonstatvar}
    k(x,x'\mid\theta,\sigma(\cdot)) = \sigma(x)\sigma(x')k^{stat}(x,x'\mid \theta)
\end{equation}
where the base kernel $k^{stat}$ is any stationary covariance function. $\sigma(x)$ represents the local standard deviation of $f(x)$. Since $k^{stat}$ is a valid covariance function and $\sigma(\cdot)>0$ the function is clearly positive-definite. The corresponding covariance function under our framework is 
\begin{equation}
\label{eq:newnonstatvar}
    k(x,x'\mid \theta, w) = g(x\mid w)g(x'\mid w)k^{stat}(x,x'\mid\theta)
\end{equation}
where $g$ is a neural network with positive output of dimension one. The positivity constraint can be enforced by using a positive activation function in the output layer of the neural network, such as exponential or softplus. 

Our framework can also be applied to covariance functions with multiple nonstationary parameters. In the one-dimensional nonstationary variance and lengthscale covariance function 
\begin{equation}
\label{eq:nonstatvarls}
    k(x,x'\mid \theta,\sigma(\cdot),\ell(\cdot)) = \sigma(x)\sigma(x')\sqrt{\frac{2\ell(x)\ell(x')}{\ell(x)^2+\ell(x')^2}}\times k^{iso}\left(\frac{( x-x')^2}{\sqrt{\ell(x)^2+\ell(x')^2}}\middle|\ \theta\right)
\end{equation}
$\ell(\cdot)$ represents the varying lengthscale and the base kernel $k^{iso}$ is any isotropic covariance function \cite{heinonen16}. In our framework this kernel would be adapted by using a neural network with a positive output of dimension two, representing the nonstationary parameters $\Tilde{\theta}(\cdot) = (\sigma(\cdot),\ell(\cdot))$. This allows for dependencies between the two parameters to be captured.

The noise can also be modeled as a nonstationary parameter by setting $\tau_i^2 = \tau(x_i)^2$.
We then use a neural network to model $\Tilde{\theta}(\cdot)=(\sigma(\cdot),\tau(\cdot))$. If the variance of the noise does not vary, we simply set $\tau_i^2 = \tau^2$ and model $\tau^2$ as a stationary parameter.

\subsection{Training and Inference}

The neural network and GP constitute a single model and are trained jointly. To evaluate the likelihood of the model the features $x$ are first fed through the neural network, producing estimates for the nonstationary parameter at each input value. These estimates are then used to calculate the full likelihood. The gradient of the log-likelihood with respect to the neural network parameter can be calculated using the chain rule, and the parameter estimates $\hat{\theta}$ and $\hat{w}$ are found by using this gradient to maximize the likelihood. The estimates can then be used to derive posterior predictive means and variances as specified in Equations \ref{eq:predictivemean} and \ref{eq:predictivevar}. An estimate of the nonstationary parameter for a particular input $x^*$ is found by simply evaluating $g(x^*\mid \hat{w})$, and does not involve GP inference.

Since evaluating the likelihood of a GP exactly requires $O(n^3)$ time and $O(n^2)$ space, approximations are frequently used to make fitting the model feasible. We use an inducing point method that assumes the covariance matrix is low rank \cite{quinonero05}. We believe our method is compatible with other approximation methods, such as Vecchia approximation \cite{vecchia88}, but do not investigate the choice of approximation in this paper.

\subsection{Interpretation}

Our framework is equivalent to learning a basis function expansion of a transformed parameter. A neural network of depth $d$ is a composition of $d$ nonlinear functions $g = g_d(g_{d-1}(\ldots\mid w_{d-1})\mid w_d)$. Each layer $g_i:\mathbb{R}^{m_{i}}\to\mathbb{R}^{m_{i+1}}$ computes $m_{i+1}$ linear combinations of the $m_i$ inputs and applies an activation function element-wise. The final layer of the neural network $g_d:\mathbb{R}^{m_d}\to \Tilde{\Theta}$ can be interpreted as taking a linear combination of $m_d$ learned basis functions and then applying a link function. If $\dim\Tilde{\Theta}>1$, as is this case in Equation \ref{eq:nonstatvarls}, then the same set of basis functions are learned for all parameters but the estimated coefficients differ.

Note that $g$ does not have to be a deep neural network. We found success with simple neural network architectures, and present the results in Section \ref{sec:experiments}. If we let $g$ be a linear function, then we are modeling the nonstationary parameter as a linear combination of the features. This is equivalent to using the features as basis functions in a basis function expansion of the nonstationary parameter process. We consider this method in addition to the neural network based approach as we believe it may be sufficient when modeling some datasets with a large number of features. %This method is similar to some of the spatial method discussed previously, but differs in both the setting and that we are not generating basis functions.

\section{Experiments}
\label{sec:experiments}

We implement and evaluate our method on several datasets. Two nonstationary models are considered, the nonstationary variance model with the kernel in Equation \ref{eq:newnonstatvar}, and the nonstationary variance and noise model. In both cases we use a Matérn kernel with smoothness $\nu = 0.5$ as the base kernel. For $g$ we perform an architecture search, trying both a shallow neural network, consisting of a single hidden layer of fifty neurons with ReLU activations, and a simple linear combination of the features. We then select the model with the better performance on a validation set. For one dataset, a low-dimensional spatial dataset, we only use a deeper neural network with two hidden layers of fifty neurons.

We compare the performance of our framework to three baselines. The first is a stationary GP model that uses a Matérn kernel with smoothness $\nu = 0.5$. The other two are hierarchical models, one of which has nonstationary noise and variance and the other only with nonstationary variance as in Equation \ref{eq:nonstatvar}, with GP priors placed on the nonstationary parameters. The priors also use a Matérn kernel with smoothness $\nu = 0.5$. Since MCMC methods are prohibitively expensive for large datasets, we use doubly stochastic variational inference with mini-batching \cite{salimbeni17} to train the hierarchical models. All models assume constant mean.

Model performance is evaluated with mean square error (MSE), marginal log-score, and runtime. The log-score is defined as the sum of the negative log-densities of the test observations under the posterior predictive distribution \cite{katzfuss14}. Unlike MSE, the log-score takes into account the predictive uncertainty. An accurate prediction with a narrow confidence interval has a lower log-score than one with a wide one, and similarly an inaccurate observation with a wide confidence interval has a lower log-score than one with a narrow one. A lower log-score indicates better performance.

For computational efficiency we assume the covariance matrix has low-rank structure and use 100 inducing points for the stationary and neural network models. The variational inference method used 100 inducing points for both the likelihood distribution and the prior.

All methods were trained using adaptive gradient descent \cite{adam} with hyperparameter tuning performed with a grid search over three step sizes. The stationary model and both proposed models were trained for 10,000 iterations, while the hierarchical model was trained for 200 epochs with batches of 1024 observations. All implementations were run on a Nvidia A6000 GPU, with runtimes including both training and evaluating the model. The step size and number of training iterations were selected based on performance on a validation set. For the proposed method, the validation set was also used to select the form of $g$. Ten percent of each dataset was reserved as a test set, with twenty percent of the remaining reserved as a validation set. This results in a 72\% training, 18\% validation, and 10\% test set partition. For the UCI datasets, results were averaged over five such partitions.

All methods in this paper were implemented using the GPyTorch\footnote{Used under MIT License: https://gpytorch.ai/} library \cite{gardner18}, which leverages a custom Krylov subspace algorithm to compute the likelihood and score function of a GP without relying on Cholesky decompositions. GPyTorch supports a wide range of GP models, including inducing point approximations, warping methods, and variational approaches. Our framework was implemented by defining new kernel classes within GPyTorch. Since GPyTorch relies on PyTorch's automatic differentiation engine \cite{pytorch}, we were able to efficiently compute gradients when fitting our models.

All code is available online at \hyperlink{https://github.com/zjames12/NeuralParam}{https://github.com/zjames12/NeuralParam}

\subsection{UCI Datasets}

The UCI Machine Learning Repository\footnote{Used under CC-BY 4.0: https://archive.ics.uci.edu/} provides benchmark datasets from a wide range of sources. We use twelve datasets that have up to 53,437 observations and 385 features. The dimensions of each dataset are found in the Appendix. We preprocess each dataset by removing duplicate rows. The log-score and MSE of all five models are presented in Tables \ref{tab:rmse} and \ref{tab:logscore}, while the runtimes are in Table \ref{tab:times}. The associated standard errors can be found in the Appendix.

The neural network method with nonstationary noise and variance (NNP noise+var) has lower MSE and log-score than the hierarchical noise and variance model (HM noise+var) and stationary model on nine of the twelve datasets. We see similar results when comparing the neural network model with nonstationary variance (NNP var) to the corresponding hierarchical model (HM var). Overall, our framework report the best MSE on nine of the datasets and the best log-score on eight. Outperforming the stationary model indicates that introducing nonstationary noise and variance produces a better fit than keeping them fixed. Additionally, outperforming the hierarchical model indicates that our framework is competitive with other methods for fitting GP with nonstationary kernels to large datasets.

When comparing NNP noise+var to NNP var we find that leaving the noise fixed results in better fits for some datasets. We attribute this to certain datasets exhibiting either very little or stationary noise, so assuming nonstationary noise can actually produce a worse fit.

Our method requires fitting twice as many models as the stationary or hierarchical models during model tuning, since we are searching across two possible forms for $g$. Despite this, its runtime is comparable to the hierarchical model, and it is faster on several of the larger datasets. While using the neural network generally outperformed the linear combination, we found that the architecture search resulted in a better model.

\begin{table}
  \caption{MSE of two nonstationary models under our framework (NNP var, NNP var+noise), two hierarchical models fit with variational inference (HM var, HM var+noise), and a stationary model when tested on several UCI datasets. The datasets were partitioned into 72\% training, 18\% validation, and 10\% test sets with results averaged over five such partitions. Standard errors are provided in the Appendix. The lowest MSE for each dataset is in bold. All GP, including the priors in the HM models, use the Matérn with $\nu=0.5$ as the base kernel, and assume a low-rank covariance with 100 inducing points.}
  \label{tab:rmse}
  \centering
  %\begin{tabular}{llllllll}
  \begin{tabular}{lrrrrrrr}
    \toprule
    Dataset & {Stationary} & {HM var} & {HM var+noise} & {NNP var} & {NNP var+noise} \\
    \midrule
    Gas$\times10^7$         & 9.82 & 37.9 & 2.81 & 6.56 & \textbf{0.71} \\
    Skillcraft$\times10^6$  & \textbf{7.03} & 7.36 & 7.33 & 7.09 & 7.05 \\
    SML                     & 21.10 & 11.20 & 11.20 & 6.77 & \textbf{6.16} \\
    Parkinsons$\times10^1$  & 5.70 & 3.64 & 3.55 & \textbf{1.49} & 1.58 \\
    % Pumadyn                 & 1.002699 & 1.003 & \textbf{1.002695} & 1.004 & 1.0028 \\
    Pumadyn                 & 1.00 & 1.00 & \textbf{1.00} & 1.00 & 1.00 \\
    PolTele                 & 5.86 & 4.31 & \textbf{4.17} & 4.73 & 5.13 \\
    KEGG                    & 2.06 & 2.14 & 2.37 & \textbf{1.71} & 1.76 \\
    Elevators$\times10^{-6}$   & 7.18 & 7.99 & 9.09 & \textbf{4.98} & 7.16 \\
    KEGGU$\times10^{-1}$    & 1.16 & 1.54 & 1.70 & \textbf{0.28} & 0.43 \\
    Kin40k$\times10^{-1}$   & 9.52 & 9.41 & 9.41 & 9.22 & \textbf{9.10} \\
    Protein $\times10^{1}$  & 1.97 & 2.08 & 2.09 & 1.88 & \textbf{1.87} \\
    CTslice $\times10^{2}$ & 2.91 & 7.48 & 7.48 & 2.82 & \textbf{2.72} \\
    \bottomrule
  \end{tabular}
\end{table}

\begin{table}
 \caption{Log-score of two nonstationary models under our framework (NNP var, NNP var+noise), two hierarchical models fit with variational inference (HM var, HM var+noise), and a stationary model when tested on several UCI datasets. The datasets were partitioned into 72\% training, 18\% validation, and 10\% test sets with results averaged over five such partitions. Standard errors are provided in the Appendix. The lowest log-score for each dataset is in bold. All GP, including the priors in the HM models, use the Matérn with $\nu=0.5$ as the base kernel, and assume a low-rank covariance with 100 inducing points.}
  \label{tab:logscore}
  \centering
  %\begin{tabular}{llllll}
   \begin{tabular}{lrrrrr}
    \toprule
    Dataset & Stationary & HM var & HM var+noise & NNP var & NNP var+noise \\
    \midrule
    Gas$\times10^4$         & 1080 & 37600 & \textbf{0.23} & 6.76 & 1.20 \\
    Skillcraft$\times10^4$  & 8330 & \textbf{0.57} & 1.19 & 1.56 & 1.13 \\
    SML$\times10^2$         & 11.0 & 9.62 & 9.63 & 8.88 & \textbf{8.57} \\
    Parkinsons$\times10^3$  & 1.83 & 1.70 & 1.70 & \textbf{1.50} & 1.53 \\
    Pumadyn$\times10^3$     & \textbf{1.05} & 1.05 & 1.05 & 1.05 & 1.05 \\
    PolTele$\times10^3$     & 3.29 & 3.06 & 3.04 & \textbf{2.79} & 2.99 \\
    KEGG$\times10^3$        & 2.68 & 2.67 & 2.68 & \textbf{2.41} & 2.42 \\
    Elevators$\times10^3$   & -5.43 & -4.87 & -4.86 & \textbf{-5.47} & -5.46 \\
    KEGGU$\times10^2$       & -2.80 & -1.91 & 3.42 & \textbf{-16.00} & -14.60 \\
    Kin40k$\times10^3$      & 5.03 & 5.00 & 5.00 & 4.97 & \textbf{4.95} \\
    Protein$\times10^4$     & 1.15 & 1.17 & 1.17 & 1.15 & \textbf{1.14} \\
    CTslice$\times10^4$     & \textbf{2.13} & 4.58 & 4.60 & 2.27 & 2.27 \\
    \bottomrule
  \end{tabular}
\end{table}

\begin{table}
  \caption{Time in seconds of two nonstationary models under our framework (NNP var, NNP var+noise), two hierarchical models fit with variational inference (HM var, HM var+noise), and a stationary model when tested on several UCI datasets. The datasets were partitioned into 72\% training, 18\% validation, and 10\% test sets with results averaged over five such partitions. Standard errors are provided in the Appendix. All GP, including the priors in the HM models, use the Matérn with $\nu=0.5$ as the base kernel, and assume a low-rank covariance with 100 inducing points.  Models were run on a Nvidia A6000 GPU, with times including training and inference.}
  \label{tab:times}
  \centering
%  \begin{tabular}{llllll}
  \begin{tabular}{lrrrrr}
    \toprule
    Dataset & Stat. & HM var & HM var+noise & NNP var & NNP var+noise \\
    \midrule
    Gas  & 133 & 107 & 159 & 845 & 739 \\
    Skillcraft & 228 & 167 & 229 & 1004 & 1240 \\
    SML  & 221 & 168 & 567 & 870 & 716 \\
    Parkinsons  & 234 & 391 & 404 & 987 & 840 \\
    Pumadyn     & 343 & 363 & 1370 & 1042 & 761 \\
    PolTele     & 142 & 914 & 1148 & 1198 & 706 \\
    KEGG        & 156 & 748 & 1971 & 1099 & 1358 \\
    Elevators   & 263 & 859 & 1950 & 1373 & 1521 \\
    KEGGU       & 227 & 948 & 2055 & 738 & 1387 \\
    Kin40k      & 155 & 2717 & 6850 & 806 & 1067 \\
    Protein     & 252 & 5585 & 3141 & 1256 & 1113 \\
    CTslice     & 398 & 3292 & 8008 & 3218 & 4369 \\
    \bottomrule
  \end{tabular}
\end{table}

\subsection{Spatial Dataset}

\begin{figure}
  \centering
  \includegraphics[width=.5\linewidth]{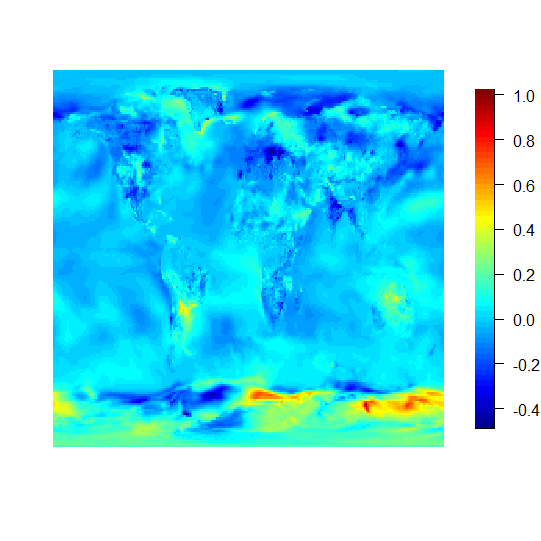}
  \caption{A pattern scaling field spatial from \cite{gerber2021}. Values represent the simulated change in local mean surface temperatures given a one degree increase in global mean surface temperatures.}
  \label{fig:climate}
\end{figure}

We test our framework on a climate model output from the NCAR Large Ensemble Project, which has been previously analyzed \cite{gerber2021}. We focus on a pattern scaling field that describes the change in local mean surface temperatures given a one-degree increase in global mean temperature. The field consists of $n=55,296$ observations and exhibits clear nonstationary behavior at low and high latitudes, as observed in Figure \ref{fig:climate}.

The dataset is partitioned into 72\% training, 18\%  validation, and 10\% test sets. All five of the models are fit to the data. Since the data is low-dimensional we expect learning the mapping $g$ to be difficult, and therefore use a deeper neural network consisting of two hidden layers with fifty neurons in each layer. Local variance estimates are produced by the neural network, while variances from the hierarchical model are sampled from the variational distribution of the prior process. We use twenty samples at each location and compute the empirical mean.

\begin{figure}[t]
  \centering
  \begin{minipage}[b]{0.49\textwidth}
    \centering
    \includegraphics[width=\linewidth]{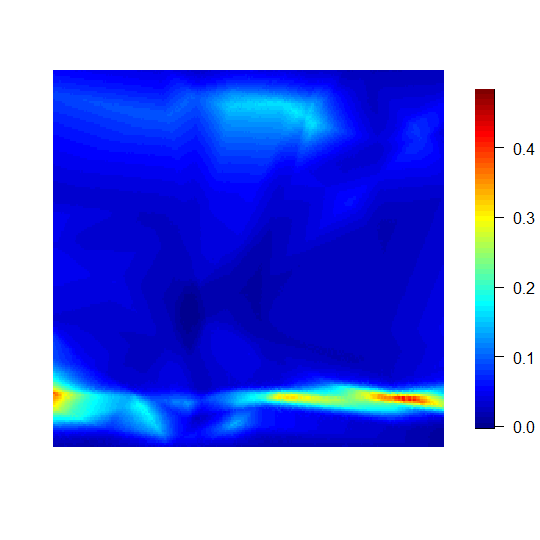}
    % \caption*{(b)}
  \end{minipage}
  \hfill
  \begin{minipage}[b]{0.49\textwidth}
    \centering
    \includegraphics[width=\linewidth]{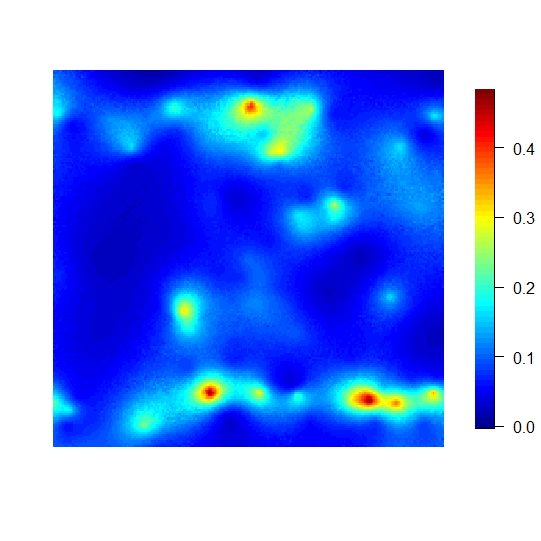}
    % \caption*{(c)}
  \end{minipage}
  \caption{Left: Local standard deviations of the pattern scaling field as estimated by the NNP method, obtained by passing locations through the neural network. Right: Local standard deviations estimated by the HM method, computed as the empirical mean of samples drawn from the prior.}
  \label{fig:climatevar}
\end{figure}

The MSE, log-score, and time of each model are reported in Table \ref{tab:climate}. The results show that the neural network with nonstationary variance outperformed the others in terms of log-score. It reports the second-best MSE, with the hierarchical variance and noise model being more accurate. The nonstationary noise models both produced noise estimates that were roughly constant across the feature space, indication that noise is stationary. The estimated local standard deviations from HM var and NNP var are in Figure \ref{fig:climatevar}, with the neural network method successfully identifying the regions of high variance. The hierarchical model also identified the regions of high variance but appears to be more sensitive to local changes. While more computationally expensive than the stationary model, the neural network model used a fraction of the time of the hierarchical model. 

\begin{table}

  \caption{Log-score, MSE, and runtime of two nonstationary models under our framework (NNP var, NNP var+noise), two hierarchical models fit with variational inference (HM var, HM var+noise), and a stationary model when fit on a spatial dataset. The dataset was partitioned into into 72\% training, 18\% validation, and 10\% test sets. All GP, including the prior in the HM models, use the Matérn with $\nu=0.5$ as the base kernel, and assume a low-rank covariance with 100 inducing points.}
  \label{tab:climate}
  \centering
  \begin{tabular}{llll}
    \toprule
    % \multicolumn{2}{c}{Part}                   \\
    % \cmidrule(r){1-2}
    Model     & MSE$\times 10^{-3}$ & Log-score$\times 10^{3}$ & Time (s) \\
    \midrule
    Stationary & 3.29  & -7.08 & 319    \\
    HM var &    2.85       & -7.52 & 4305 \\
    HM var+noise & 2.72 & -7.60 & 8104\\
    NNP var & 2.81 & -7.68 &887      \\
    NNP var+noise &3.04 & -7.48 & 1231\\
    
    \bottomrule
  \end{tabular}
\end{table}

\section{Discussion}

\paragraph{Limitations}

Our model uses the features to directly model the nonstationary parameters, which may prove inadequate for low-dimensional data. Further research is needed to evaluate our model's performance on temporal and spatial datasets. Furthermore, since we rely on neural networks, the relationship between the features and nonstationary parameters is not explicitly interpretable. As with all neural networks, predictions on points far away from the training data may be poor. Neural networks also tend to perform best on large datasets, so our method may not be appropriate on small datasets. Lastly, our method models the nonstationary parameters deterministically, limiting our ability to provide uncertainty quantification for the nonstationary parameter estimates.

Our paper presents a new framework for modeling nonstationary GPs. By modeling the nonstationary parameters in a nonstationary kernel with a neural network, we give practitioners fine control over the kernel and a high degree of interpretability. Furthermore, we do not sacrifice computational efficiency, as our framework is compatible with approximation methods. Future research will explore more complex neural network architectures and assess how the method performs in low-dimensional spaces, where the features may not contain sufficient information to model the parameters of interest. We hope our framework enables practitioners to efficiently fit a wide range of nonstationary GP models without being hindered by model setup and training.

\bibliographystyle{unsrt}
\bibliography{bib}

%%%%%%%%%%%%%%%%%%%%%%%%%%%%%%%%%%%%%%%%%%%%%%%%%%%%%%%%%%%%
\clearpage{}\newpage{}
\appendix
\section{Appendix}
\begin{table*}[!thp]
  \caption{UCI Datasets with number of observations $n$ and features $p$. Duplicate rows are removed.}
  \label{sample-table}
  \centering
  \begin{tabular}{lll}
    \toprule
    Dataset & $n$ & $p$ \\
    \midrule
    Gas  & 2570 & 128 \\
    Skillcraft         & 3340 & 19 \\
    SML  & 4137 & 26 \\
    Parkinsons         & 5875 & 20 \\
    Pumadyn     & 8192 & 32 \\
    PolTele     & 14958 & 26 \\
    KEGG        & 16341 & 22 \\
    Elevators   & 16599 & 18 \\
    KEGGU       & 18575 & 27 \\
    Kin40k      & 40000 & 8 \\
    Protein     & 44020 & 9 \\
    CTslice     & 53437 & 385 \\
    \bottomrule
  \end{tabular}
\end{table*}

\begin{table*}[!thp]
  \caption{Standard errors of the MSE of two nonstationary models under our framework (NNP var, NNP var+noise), two hierarchical models fit with variational inference (HM var, HM var+noise), and a stationary model when tested on several UCI datasets. Standard errors are calculated under a normality assumption. The datasets were partitioned into 72\% training, 18\% validation, and 10\% test sets with results averaged over five such partitions. All GP, including the priors in the HM models, use the Matérn with $\nu=0.5$ as the base kernel, and assume a low-rank covariance with 100 inducing points}
  \label{tab:msese}
  \centering
  \begin{tabular}{llllll}
    \toprule
    Dataset & Stationary & HM var & HM var+noise & NNP var & NNP var+noise \\
    \midrule
    Gas$\times10^7$& 3.60 & 35.3 & 1.15 & 0.994 & 0.022 \\
    Skillcraft $\times10^4$ & 8.95 & 6.72 & 13.4 & 9.43 & 24.8 \\
    SML         & 1.40 & 0.746 & 1.63 & 0.929 & 2.23 \\
    Parkinsons  & 2.36 & 1.62 & 5.33 & 1.10 & 2.50 \\
    Pumadyn$\times10^{-2}$     & 1.37 & 1.37 & 3.06 & 1.53 & 3.03 \\
    PolTele     & 0.340 & 0.301 & 1.02 & 1.07 & 1.39 \\
    KEGG $\times10^{-1}$& 1.39 & 2.09 & 5.95 & 1.70 & 4.07 \\
    Elevators$\times10^{-7}$   & 4.21 & 5.08 & 28.0 & 2.10 & 48.0 \\
    KEGGU$\times10^{-2}$& 8.54 & 13.4 & 30.6 & 1.89 & 7.57 \\
    Kin40k$\times10^{-3}$      & 3.95 & 2.95 & 6.84 & 6.32 & 14.0 \\
    Protein$\times10^{-7}$& 3.61 & 3.06 & 7.07 & 3.52 & 6.11 \\
    CTslice     & 1.54 & 4.20 & 9.39 & 11.9 & 18.9 \\
    \bottomrule
  \end{tabular}
\end{table*}

\begin{table*}[!thp]
  \caption{Standard errors of the log-score of two nonstationary models under our framework (NNP var, NNP var+noise), two hierarchical models fit with variational inference (HM var, HM var+noise), and a stationary model when tested on several UCI datasets. Standard errors are calculated under a normality assumption. The datasets were partitioned into 72\% training, 18\% validation, and 10\% test sets with results averaged over five such partitions. All GP, including the priors in the HM models, use the Matérn with $\nu=0.5$ as the base kernel, and assume a low-rank covariance with 100 inducing points}
  \label{tab:logscorese}
  \centering
  \begin{tabular}{llllll}
    \toprule
    Dataset & Stat. & HM var & HM var+noise & NNP var & NNP var+noise \\
    \midrule
    Skillcraft$\times10^{3}$  & 1510 & 0.129 & 6.29 & 3.95 & 4.28 \\
    Gas $\times10^{4}$        & 355 & 37600 & 0.005 & 1.08 & 0.27 \\
    SML $\times10^{1}$        & 1.44 & 1.25 & 1.31 & 1.47 & 1.63 \\
    Parkinsons$\times10^{1}$  & 1.22 & 1.53 & 2.51 & 2.96 & 3.75 \\
    Pumadyn                   & 0.007 & 4.56 & 4.69 & 5.37 & 4.70 \\
    PolTele$\times10^{1}$     & 1.20 & 0.78 & 1.50 & 21.7 & 1.67 \\
    KEGG$\times10^{1}$        & 1.61 & 1.59 & 1.28 & 2.39 & 2.35 \\
    Elevators                 & 2.90 & 1.38 & 2.20 & 1.29 & 2.08 \\
    KEGGU $\times10^{2}$      & 1.59 & 2.30 & 2.33 & 0.101 & 1.22 \\
    Kin40k                    & 6.52 & 5.46 & 5.64 & 11.9 & 13.2 \\
    Protein$\times10^{1}$     & 3.80 & 3.62 & 3.69 & 4.31 & 3.62 \\
    CTslice$\times10^{2}$     & 0.28 & 1.73 & 1.74 & 2.64 & 1.81 \\
    \bottomrule
  \end{tabular}
\end{table*}

\begin{table*}[!thp]
  \caption{Standard errors of the times of two nonstationary models under our framework (NNP var, NNP var+noise), two hierarchical models fit with variational inference (HM var, HM var+noise), and a stationary model when tested on several UCI datasets. Standard errors are calculated under a normality assumption. The datasets were partitioned into 72\% training, 18\% validation, and 10\% test sets with results averaged over five such partitions. All GP, including the priors in the HM models, use the Matérn with $\nu=0.5$ as the base kernel, and assume a low-rank covariance with 100 inducing points}
  \label{tab:timese}
  \centering
  \begin{tabular}{llllll}
    \toprule
    Dataset & Stat. & HM var & HM var+noise & NNP var & NNP var+noise \\
    \midrule
    Gas         & 1.16 & 2.94 & 0.34 & 15.8 & 7.65 \\
    Skillcraft  & 0.658 & 2.78 & 1.19 & 11.0 & 12.0 \\
    SML         & 2.52 & 1.03 & 0.461 & 12.2 & 24.6 \\
    Parkinsons  & 3.47 & 0.473 & 0.825 & 18.8 & 9.59 \\
    Pumadyn     & 0.00739 & 1.36 & 2.35 & 17.1 & 8.30 \\
    PolTele     & 0.795 & 0.453 & 9.61 & 19.8 & 8.90 \\
    KEGG        & 1.37 & 1.47 & 2.43 & 28.7 & 23.7 \\
    Elevators   & 6.09 & 0.766 & 3.30 & 9.24 & 13.7 \\
    KEGGU       & 1.99 & 10.2 & 4.74 & 19.7 & 80.7 \\
    Kin40k      & 1.49 & 5.92 & 367 & 17.0 & 18.6 \\
    Protein     & 3.16 & 21.9 & 3.19 & 9.04 & 25.2 \\
    CTslice     & 9.39 & 10.5 & 55.0 & 235 & 45.8 \\
    \bottomrule
  \end{tabular}
\end{table*}

\end{document}